\documentclass[conference]{ieeeconf}
\usepackage[left=1.91cm,right=1.91cm,top=1.91cm,bottom=1.91cm]{geometry}
\pdfoutput=1

\IEEEoverridecommandlockouts                              

\usepackage[utf8]{inputenc}
\usepackage{graphicx}

\usepackage{amsmath} 
\usepackage{amsthm}  
\usepackage{amsfonts}
\usepackage{breqn}
\usepackage{dsfont}
\usepackage[dvipsnames]{xcolor}
\usepackage[nolist,nohyperlinks]{acronym}
\usepackage[linesnumbered,ruled,vlined]{algorithm2e}
\usepackage[capitalize]{cleveref}
\Crefname{equation}{}{}
\usepackage[per-mode=symbol]{siunitx}
\usepackage{tikz}
\usepackage{pgfplots}\pgfplotsset{compat=1.16}\newlength{\figurewidth}\newlength{\figureheight}
\usepackage{subcaption}
\usepackage{csquotes}

\usepackage[utf8]{inputenc}   				 	
\usepackage{silence}  							
\usepackage{url}
\usepackage{letltxmacro}
\LetLtxMacro{\autocite}{\cite}
\LetLtxMacro{\textcite}{\cite}

\title{Constrained Sampling from a Kernel Density Estimator to Generate Scenarios for the Assessment of Automated Vehicles}
\author{Erwin de Gelder$^{1,2*}$, Eric Cator$^{3}$, Jan-Pieter Paardekooper$^{1,4}$, Olaf Op den Camp$^{1}$, Bart De Schutter$^{2}$%
\thanks{$^{1}$TNO, Integrated Vehicle Safety, Helmond, The Netherlands}%
\thanks{$^{2}$Delft University of Technology, Delft Center for Systems and Control, Delft, The Netherlands}%
\thanks{$^{3}$Radboud University, Applied Stochastics, Nijmegen, The Netherlands}%
\thanks{$^{4}$Radboud University, Donders Institute for Brain, Cognition and Behaviour, Nijmegen, The Netherlands}%
\thanks{$^{*}$Corresponding author. \newline E-mail address: {\tt\small erwin.degelder@tno.nl}}}%
\date{}

\newcommand{\accelerationsymbol}{a}
  \newcommand{\accelerationinit}{\accelerationsymbol_{\mathrm{init}}}
\newcommand{\bandwidth}{h}
\newcommand{\bandwidthmatrix}{H}

  \newcommand{\bandwidthmatrrotatedixinverse}{\Lambda}
  \newcommand{\bwriul}{\bandwidthmatrrotatedixinverse_{11}}
  \newcommand{\bwriur}{\bandwidthmatrrotatedixinverse_{12}}
  \newcommand{\bwribl}{\bandwidthmatrrotatedixinverse_{21}}
  \newcommand{\bwribr}{\bandwidthmatrrotatedixinverse_{22}}
  \newcommand{\bwrischur}{\bandwidthmatrrotatedixinverse_{\mathrm{S}}}
\newcommand{\constantterm}{C}
\newcommand{\constraintmatrix}{A}
\newcommand{\constraintvector}{b}
\newcommand{\density}[1]{f\left( #1 \right)}
\newcommand{\densityest}[1]{\hat{f}\left( #1 \right)}

\newcommand{\determinant}[1]{\left| #1 \right|}
\newcommand{\dimension}{d}
  \newcommand{\dimensionparta}{\dimension_\mathrm{c}}
  \newcommand{\dimensionpartb}{\dimension_\mathrm{u}}
\newcommand{\dummyvar}{u}
\newcommand{\e}[1]{\exp\left\{ #1 \right\}}
\newcommand{\identitymatrix}[1]{I_{#1}}
\newcommand{\indexdata}{i}
\newcommand{\indexsampling}{j}
\newcommand{\interquartilerange}{R}
\newcommand{\kernelfunc}[1]{K \left( #1 \right)}
\newcommand{\kernelfuncnormalized}[2]{K_{#1} \left( #2 \right)}
\newcommand{\normtwo}[1]{\left\Vert #1 \right\Vert_2}
\newcommand{\numberofconstraints}{n_\mathrm{c}}
\newcommand{\numberofsamples}{N}
\newcommand{\realnumbers}{\mathds{R}}
\newcommand{\speedsymbol}{v}
  \newcommand{\speed}[1]{\speedsymbol\left(#1\right)}
  \newcommand{\speedinit}{\speedsymbol_{\mathrm{init}}}
  \newcommand{\speedend}{\speedsymbol_{\mathrm{end}}}
\newcommand{\std}{\sigma}
\newcommand{\superquad}{\phantom{=}\quad\quad\quad\quad}
\newcommand{\svdu}{U}
  \newcommand{\svds}{\Sigma}
  \newcommand{\svdv}{V}
  \newcommand{\svdva}{V_1}
  \newcommand{\svdvb}{V_2}
  \newcommand{\svduB}{\bar{\svdu}}
  \newcommand{\svduBa}{\svduB_1}
  \newcommand{\svduBb}{\svduB_2}
  \newcommand{\svduBvec}[1]{\bar{u}_{#1}}
  \newcommand{\svdsB}{\bar{\svds}}
  \newcommand{\svdsBa}{\svdsB_1}
  \newcommand{\svdsBb}{\svdsB_2}
  \newcommand{\svdvB}{\bar{\svdv}}
  \newcommand{\svdvBa}{\svdvB_1}
  \newcommand{\svdvBb}{\svdvB_2}
  \newcommand{\svdvBvec}[1]{\bar{v}_{#1}}
\renewcommand{\time}{t}
  \newcommand{\timestep}{\Delta \time}
  \newcommand{\timestepnumberof}{n_{\time}}
  \newcommand{\timedatapoint}[1]{\time_{#1}}
\newcommand{\transpose}{^{\mkern-1.5mu\mathsf{T}}}

\newcommand{\variable}{x}
  \newcommand{\variableconstrained}{\bar{\variable}}
  \newcommand{\variableunconstrained}{\tilde{\variable}}
  \newcommand{\datapoint}[1]{\variable_{#1}}
  \newcommand{\datapointconstrained}[1]{\variableconstrained_{#1}}
  \newcommand{\datapointunconstrained}[1]{\variableunconstrained_{#1}}
  \newcommand{\datapointunconstrainedtranslated}[1]{\datapointunconstrained{#1}'}
  \newcommand{\datapointmatrix}{X}
  \newcommand{\datapointmean}{\mu}
  \newcommand{\datapointmeanentry}[1]{\datapointmean_{#1}}
\newcommand{\weight}[1]{w_{#1}}

\newcommand{\cstarta}{}  
\newcommand{\cenda}{}
\newcommand{\cstartb}{}  
\newcommand{\cendb}{}
\newcommand{\cstartc}{}  
\newcommand{\cendc}{}

\usepackage{nopageno}

\makeatletter
\patchcmd{\@algocf@start}
  {-1.5em}
  {0pt}
  {}{}
\makeatother

\begin{document}

\begin{acronym}[AAAAAAAA]
	\acro{av}[AV]{Automated Vehicle}\acroindefinite{av}{an}{an}
	\acro{kde}[KDE]{Kernel Density Estimation}
	\acro{ngsim}[NGSIM]{Next Generation SIMulation}\acroindefinite{ngsim}{an}{a}
	\acro{pca}[PCA]{Principal Componant Analysis}
	\acro{pdf}[pdf]{probability density function}
	\acro{svd}[SVD]{Singular Value Decomposition}\acroindefinite{svd}{an}{a}
\end{acronym}

\maketitle

\begin{abstract}

The safety assessment of \acp{av} is an important aspect of the development cycle of \acp{av}.
A scenario-based assessment approach is accepted by many players in the field as part of the complete safety assessment.
\cstarta A scenario is a representation of a situation on the road to which the \ac{av} needs to respond appropriately. \cenda
One way to generate the required scenario-based test descriptions is to parameterize the scenarios and to draw these parameters from \iac{pdf}.
\cstarta Because the shape of the \ac{pdf} is unknown beforehand, assuming a functional form of the \ac{pdf} and fitting the parameters to the data may lead to inaccurate fits.
As an alternative, \ac{kde} is a promising candidate for estimating the underlying \ac{pdf}, because it is flexible with the underlying distribution of the parameters. \cenda
Drawing random samples from \iac{pdf} estimated with \ac{kde} is possible without the need of evaluating the actual \ac{pdf}, which makes it suitable for drawing random samples for, e.g., Monte Carlo methods.
\cstarta Sampling from \iac{kde} while the samples satisfy a linear equality constraint, however, has not been described in the literature, as far as the authors know. \cenda

In this paper, we propose a method to sample from \iac{pdf} estimated using \ac{kde}, such that the samples satisfy a linear equality constraint. 
We also present an algorithm of our method in pseudo-code. 
\cstartb The method can be used to generating scenarios that have, e.g., a predetermined starting speed or to generate different types of scenarios.
This paper also shows that the method for sampling scenarios can be used in case \iac{svd} is used to reduce the dimension of the parameter vectors. \cendb

\end{abstract}

\acresetall

\section{Introduction}
\label{sec:introduction}

An essential facet in the development of \acp{av} is the assessment of quality and performance aspects of the \acp{av}, such as safety, comfort, and efficiency \autocite{bengler2014threedecades, stellet2015taxonomy, koopman2016challenges}. 
Because public road tests are expensive and time consuming \autocite{kalra2016driving, zhao2018evaluation}, a scenario-based approach has been proposed \autocite{stellet2015taxonomy, riedmaier2020survey, elrofai2018scenario, putz2017pegasus, deGelder2017assessment, jacobo2019development}.
As a source of information for the scenarios for the assessment, real-world driving data has been proposed, such that the scenarios relate to real-world driving conditions \autocite{elrofai2018scenario, putz2017pegasus, krajewski2018highD}.

When using scenarios extracted from real-world driving data as a direct source for describing scenario-based tests, two problems arise.
First, not all possible variations of the scenarios might be found in the data. 
Therefore, the failure modes of the \acp{av} might not be reflected in the tests that are based on the scenarios that are extracted from real-world driving data \autocite{zhao2018evaluation}.
Second, using scenarios extracted from real-world driving data might not reduce the actual testing load, because the set of extracted scenarios is largely composed of non-safety critical scenarios \autocite{zhao2018evaluation}.
As a solution to this, so-called importance sampling has been introduced in order to put more emphasis on scenarios that are likely to trigger safety-critical situations \autocite{zhao2018evaluation, deGelder2017assessment, xu2018accelerated, jesenski2020scalable}.
These methods \autocite{zhao2018evaluation, deGelder2017assessment, xu2018accelerated, jesenski2020scalable} have in common that they describe scenarios using parameters for which \iac{pdf} is estimated.

As already presented in \autocite{deGelder2017assessment, degelder2019risk}, we propose to estimate the \ac{pdf} using \ac{kde}.
\ac{kde} \autocite{parzen1962estimation, rosenblatt1956remarks} is often referred to as a non-parametric way to estimate the \ac{pdf}, because no use is made of a predefined functional form of the \ac{pdf} for which certain parameters are fitted to the data. 
\cstarta Because \ac{kde} produces \iac{pdf} that adapts itself to the data, it is flexible regarding the shape of the actual underlying distribution of the parameters. \cenda

Sampling from \iac{kde} is straightforward.
In some cases, however, one wants to sample from the estimated \ac{pdf} while a part of the random sample is fixed.
\cstartb For example, one may want to assess the performance of \iac{av} when it is driving at its maximum allowable speed. \cendb
Conditional sampling allows to generate scenario-based test cases in which, e.g., the ego vehicle has a fixed speed. 
One approach to performing conditional sampling is to evaluate the conditional \ac{pdf} and to use this for sampling.
This method, however, would be highly cumbersome, especially with a higher-dimensional \ac{pdf}\cstarta, because marginal integrals of codimension 1 of the conditional \ac{pdf} must be evaluated\cenda. 

We will propose an algorithm to sample parameters from \iac{kde} while the parameters are subject to a linear equality constraint.
\cstartb Our work differs from \autocite{hall1999density, wolters2018practical} because these works consider (shape) constraints on the estimated \ac{pdf}. 
The proposed algorithm can be regarded as a generalization of the conditional density estimation in \autocite{holmes2007fast}. \cendb
\cstarta Our proposed method is efficient, because the actual (conditional) \ac{pdf} does not need to be evaluated. \cenda
We illustrate the proposed sampling technique and its practical usefulness using an example.
Furthermore, we will explain the usefulness of sampling with linear equality constraints in case parameter reduction techniques are used to avoid the curse of dimensionality that \ac{pdf} estimation techniques, such as \ac{kde}, are suffering from.

This paper is organized as follows.
In \cref{sec:problem}, we first describe the problem in more detail.
The proposed method is presented in \cref{sec:method}.
Through an example, we illustrate the correct performance of the algorithm in \cref{sec:example}.
We also apply the proposed method to sample different types of scenarios in \cref{sec:example}.
In \cref{sec:discussion}, some implications and limitations of this work are discussed. 
Conclusions of the paper are provided in \cref{sec:conclusions}.

\section{Problem definition}
\label{sec:problem}

In \ac{kde}, the \ac{pdf} $\density{\cdot}$ is estimated as follows:
\begin{equation}
	\label{eq:kde}
	\densityest{\variable} = 
	\frac{1}{\numberofsamples} \sum_{\indexdata=1}^{\numberofsamples} 
	\kernelfuncnormalized{\bandwidthmatrix}{\variable - \datapoint{\indexdata}}.
\end{equation}
Here, $\datapoint{\indexdata} \in \realnumbers^{\dimension}$ represents the $i$-th data point of dimension $\dimension$.
In total, there are $\numberofsamples$ data points, so $i \in \{1, \ldots, \numberofsamples\}$.
In \cref{eq:kde}, $\kernelfuncnormalized{\bandwidthmatrix}{\cdot}$ is the so-called scaled kernel with a positive definite symmetric bandwidth matrix $\bandwidthmatrix \in \realnumbers^{\dimension \times \dimension}$.
The kernel $\kernelfunc{\cdot}$ and the scaled kernel $\kernelfuncnormalized{\bandwidthmatrix}{\cdot}$ are related using
\begin{equation}
	\label{eq:scaled kernel}
	\kernelfuncnormalized{\bandwidthmatrix}{\dummyvar}
	= \determinant{\bandwidthmatrix}^{-1/2} \kernelfunc{ \bandwidthmatrix^{-1/2} \dummyvar },
\end{equation}
where $\determinant{\cdot}$ denotes the matrix determinant.
The choice of the kernel function is not as important as the choice of the bandwidth matrix \autocite{turlach1993bandwidthselection, duong2007ks}.
Often, a Gaussian kernel is opted and this paper is no exception.
The Gaussian kernel is given by
\begin{equation}
	\label{eq:gaussian kernel}
	\kernelfunc{\dummyvar} = \frac{1}{\left(2 \pi \right)^{\dimension/2}} \e{ -\frac{1}{2} \normtwo{\dummyvar}^2},
\end{equation}
where $\normtwo{\dummyvar}^2=\dummyvar\transpose \dummyvar$ denotes the squared 2-norm of $\dummyvar$.
Substituting \cref{eq:gaussian kernel,eq:scaled kernel} into \cref{eq:kde} gives
\begin{equation}
	\label{eq:gaussian kde}
	\densityest{\variable}
	= \constantterm
	\sum_{\indexdata=1}^{\numberofsamples} 
	\e{-\frac{1}{2} \left( \variable - \datapoint{\indexdata} \right)\transpose H^{-1} \left( \variable - \datapoint{\indexdata} \right)},
\end{equation}
where $\constantterm=\frac{1}{N \left(2\pi\right)^{d/2} \determinant{\bandwidthmatrix}^{1/2}}$ is a constant.

The bandwidth matrix is an important parameter of the \ac{kde}. 
\cstartc Several methods have been proposed to estimate the bandwidth matrix based on the available data. 
Perhaps the simplest method is Silverman's rule of thumb \autocite{silverman1986density}, in which $\bandwidthmatrix=\bandwidth^2 \identitymatrix{\dimension}$ with $\identitymatrix{\dimension}$ denoting the $\dimension$-by-$\dimension$ identity matrix\footnote{If $\bandwidthmatrix=\bandwidth^2\identitymatrix{\dimension}$, the same smoothing is applied in every direction. Therefore, the data are often normalized before applying \ac{kde} with $\bandwidthmatrix=\bandwidth^2 \identitymatrix{\dimension}$.} and 
\begin{equation}
	\bandwidth = 1.06 \min \left(\std, \frac{\interquartilerange}{1.34}\right) \numberofsamples^{-\frac{1}{5}}.
\end{equation}
Here, $\std$ denotes the standard deviation of the data and $\interquartilerange$ is the interquartile range of the data. 
Another strategy is to use cross validation.
As a special case, with one-leave-out cross validation, the bandwidth matrix equals
\begin{equation}
	\arg \min_{\bandwidthmatrix}
	\prod_{\indexdata=1}^{\numberofsamples}
	\left(
		\frac{1}{\numberofsamples-1}
		\sum_{\indexsampling=1,\indexsampling\ne\indexdata}^{\numberofsamples}
		\kernelfuncnormalized{\bandwidthmatrix}{\datapoint{\indexdata}-\datapoint{\indexsampling}}
	\right).
\end{equation}
Selecting the bandwidth matrix by this method minimizes the Kullback-Leibler divergence between $\density{\cdot}$ and $\densityest{\cdot}$ \autocite{turlach1993bandwidthselection}.
Another often-used strategy is to use plug-in methods. 
The idea of plug-in methods is to select an initial $\bandwidthmatrix$ and then plug $\densityest{\cdot}$ into an equation that calculates the optimal\footnote{Optimal in the sense that it minimizes a specific value, which is usually the asymptotic mean integrated squared error.} bandwidth based on a given \ac{pdf}.
This process is then iterated until $\bandwidthmatrix$ converges.
\cendc
We refer the interested reader to \autocite{turlach1993bandwidthselection, duong2007ks, jones1996brief, gramacki2017fft} for details on the estimation of $\bandwidthmatrix$.
In this paper, we assume that $\bandwidthmatrix$ is given.

Sampling new data points from $\densityest{\cdot}$ of \cref{eq:gaussian kde} is straightforward.
First, an integer $\indexsampling\in\{1,\ldots,\numberofsamples\}$ is randomly chosen with each integer having equal likelihood. 
Next, a random \cstarta sample is drawn from a Gaussian with covariance $\bandwidthmatrix$ and mean $\datapoint{\indexsampling}$. \cenda

\cstartb In this paper, we want to sample from \cref{eq:gaussian kde}, such that the samples satisfy the \cendb linear equality constraint:
\begin{equation}
	\label{eq:linear constraint}
	\constraintmatrix \variable = \constraintvector.
\end{equation}
Here $\constraintmatrix \in \realnumbers^{\numberofconstraints \times \dimension}$ and $\constraintvector \in \realnumbers^{\numberofconstraints}$ denote the constraint matrix and the constraint vector, respectively. 
It is assumed that the constraint matrix $\constraintmatrix$ has full rank.
Note that if $\constraintmatrix$ has not full rank, the constraint of \cref{eq:linear constraint} can easily be reformulated using Gaussian elimination, resulting in a similar constraint with a constraint matrix that has full rank.
In total, there are $\numberofconstraints < \dimension$ constraints.

\section{Method}
\label{sec:method}

To deal with the constraint \cref{eq:linear constraint}, we will perform a rotation of $\variable$, such that a part of the resulting vector is fixed by the constraint \cref{eq:linear constraint}, while the other part of the resulting vector can be freely chosen.
To perform the rotation, we employ \iac{svd} \autocite{golub2013matrix} of $\constraintmatrix$:
\begin{equation}
	\label{eq:svd constraint matrix}
	\constraintmatrix 
	= \svdu \begin{bmatrix} \svds & 0 \end{bmatrix} \svdv\transpose
	= \svdu
	\begin{bmatrix} \svds & 0 \end{bmatrix}
	\begin{bmatrix} \svdva\transpose \\ \svdvb\transpose \end{bmatrix}
	= \svdu \svds \svdva\transpose.
\end{equation}
Here, $\svdu \in \realnumbers^{\dimensionparta \times \dimensionparta}$ and $\svdv \in \realnumbers^{\dimension \times \dimension}$ are orthonormal matrices, i.e., $\svdu^{-1} = \svdu\transpose$ and $\svdv^{-1} = \svdv\transpose$.
The first $\dimensionparta$ columns of $\svdv$ are denoted by $\svdva$ while $\svdvb$ denotes the remaining columns of $\svdv$.
Moreover, $\svds \in \realnumbers^{\dimensionparta \times \dimensionparta}$ is a diagonal matrix with its so-called singular values on its diagonal.
Because $\constraintmatrix$ has full rank and $\numberofconstraints<\dimension$, all singular values are strictly positive.
As such, evaluating $\svds^{-1}$ is straightforward.
Now, let $\variableconstrained \in \realnumbers^{\dimensionparta}$ and $\variableunconstrained \in \realnumbers^{\dimensionpartb}$ such that
\begin{equation}
	\label{eq:rotation variable}
	\variable = \svdva \variableconstrained + \svdvb \variableunconstrained 
	= \begin{bmatrix} \svdva & \svdvb \end{bmatrix} \begin{bmatrix} \variableconstrained \\ \variableunconstrained \end{bmatrix}
	= \svdv \begin{bmatrix} \variableconstrained \\ \variableunconstrained \end{bmatrix}.
\end{equation}
Note that because $\svdv^{-1} = \svdv\transpose$, we have $\variableconstrained = \svdva\transpose \variable$ and $\variableunconstrained = \svdvb\transpose \variable$.
\cstarta Moreover, $\svdva\transpose\svdva=\identitymatrix{\dimensionparta}$ and $\svdva\transpose \svdvb = 0$, such that \cenda substituting \cref{eq:svd constraint matrix,eq:rotation variable} into \cref{eq:linear constraint}, gives
\begin{equation}
	\label{eq:linear constraint after rotation}
	\svdu \svds \svdva\transpose \left( \svdva \variableconstrained + \svdvb \variableunconstrained \right)
	= \svdu \svds \variableconstrained = \constraintvector.
\end{equation}
This means that in order to satisfy the constraint \cref{eq:linear constraint}, $\variableunconstrained$ can take any value whereas $\variableconstrained$ is fixed:
\begin{equation}
	\label{eq:fixed part}
	\variableconstrained = \svds^{-1} \svdu\transpose b.
\end{equation}

Similar as $\variableconstrained$ and $\variableunconstrained$, let $\datapointconstrained{\indexdata}=\svdva\transpose \datapoint{\indexdata}$ and $\datapointunconstrained{\indexdata}=\svdvb\transpose \datapoint{\indexdata}$. 
Using this and \cref{eq:rotation variable}, we can rewrite \cref{eq:gaussian kde}:
\begin{equation}
	\label{eq:density rotation}
	\densityest{\variable} = \constantterm \sum_{\indexdata=1}^{\numberofsamples}
	\e{ -\frac{1}{2} 
		\begin{bmatrix} 
			\variableconstrained - \datapointconstrained{\indexdata} \\ 
			\variableunconstrained - \datapointunconstrained{\indexdata}
		\end{bmatrix}\transpose \svdv\transpose \bandwidthmatrix^{-1} \svdv
		\begin{bmatrix} 
			\variableconstrained - \datapointconstrained{\indexdata} \\ 
			\variableunconstrained - \datapointunconstrained{\indexdata}
		\end{bmatrix}
	}.
\end{equation}
\cstartb To ease the notation, \cendb let us use the following notation:
\begin{equation}
	\svdv\transpose \bandwidthmatrix^{-1} \svdv 
	= \begin{bmatrix} \bwriul & \bwriur \\ \bwribl & \bwribr \end{bmatrix}, \label{eq:rotated matrix inverse}
\end{equation}
with $\bwriul \in \realnumbers^{\dimensionparta \times \dimensionparta}$, $\bwriur \in \realnumbers^{\dimensionparta \times \dimensionpartb}$, $\bwribl \in \realnumbers^{\dimensionpartb \times \dimensionparta}$, and $\bwribr \in \realnumbers^{\dimensionpartb \times \dimensionpartb}$. 
\cstartb Using the Schur complement \autocite{zhang2006schur} \cendb
\begin{equation}
	\bwrischur = \bwriul - \bwriur \bwribr^{-1} \bwribl, \label{eq:rotated schur complement}
\end{equation}
\cstartb we can write \cref{eq:rotated schur complement} as
\begin{equation*}
	\svdv\transpose \bandwidthmatrix^{-1} \svdv =
	\begin{bmatrix} \identitymatrix{\dimensionparta} & \bwriur \bwribr^{-1} \\ 0 & \identitymatrix{\dimensionpartb} \end{bmatrix}
	\begin{bmatrix} \bwrischur  & 0 \\ 0 & \bwribr \end{bmatrix}
	\begin{bmatrix} \identitymatrix{\dimensionparta}  & 0 \\ \bwribr^{-1} \bwribl & \identitymatrix{\dimensionpartb} \end{bmatrix}.
\end{equation*}
Substituting this in the exponent of \cref{eq:density rotation} gives
\begin{align*}
	&\begin{bmatrix} 
		\variableconstrained - \datapointconstrained{\indexdata} \\ 
		\variableunconstrained - \datapointunconstrained{\indexdata} 
	\end{bmatrix}\transpose 
	\bandwidthmatrix^{-1}
	\begin{bmatrix} 
		\variableconstrained - \datapointconstrained{\indexdata} \\ 
		\variableunconstrained - \datapointunconstrained{\indexdata} 
	\end{bmatrix} \\
	&= \begin{bmatrix} 
		\variableconstrained - \datapointconstrained{\indexdata} \\ 
		\variableunconstrained - \datapointunconstrained{\indexdata} + \bwribr^{-1} \bwriur\transpose \left(\variableconstrained - \datapointconstrained{\indexdata} \right)
	\end{bmatrix}\transpose 
	\begin{bmatrix} 
		\bwrischur  & 0 \\ 
		0 & \bwribr 
	\end{bmatrix} \\
	&\superquad \begin{bmatrix}
		\variableconstrained - \datapointconstrained{\indexsampling} \\
		\variableunconstrained - \datapointunconstrained{\indexsampling} + \bwribr^{-1}\bwribl \left(\variableconstrained - \datapointconstrained{\indexsampling} \right)
	\end{bmatrix} \\
	&= \left(\variableconstrained - \datapointconstrained{\indexdata}\right)\transpose 
	\bwrischur 
	\left(\variableconstrained - \datapointconstrained{\indexdata}\right) + \nonumber \\
	&\superquad \left( \variableunconstrained - \datapointunconstrained{\indexdata} + \bwribr^{-1} \bwriur\transpose \left(\variableconstrained - \datapointconstrained{\indexdata} \right) \right)\transpose
	\bwribr \\
	&\superquad \left( \variableunconstrained - \datapointunconstrained{\indexdata} + \bwribr^{-1} \bwriur\transpose \left(\variableconstrained - \datapointconstrained{\indexdata} \right) \right).
\end{align*}
Using this, \cref{eq:density rotation} can be written as \cendb
\begin{equation*}
	\densityest{\variable}
	= \constantterm \sum_{\indexdata=1}^{\numberofsamples} \weight{\indexdata}
	\e{ -\frac{1}{2} \left( \variableunconstrained - \datapointunconstrainedtranslated{\indexdata} \right)\transpose
		\bwribr \left( \variableunconstrained - \datapointunconstrainedtranslated{\indexdata} \right)},
\end{equation*}
with
\begin{align}
	\label{eq:weights constrained hard}
	\weight{\indexdata} &= \e{ -\frac{1}{2} \left( \variableconstrained - \datapointconstrained{\indexdata} \right)\transpose
		\bwrischur
		\left( \variableconstrained - \datapointconstrained{\indexdata} \right)},
	\forall \indexdata \in \{1,\ldots,\numberofsamples\}, \\
	\label{eq:offset constrained hard}
	\datapointunconstrainedtranslated{\indexdata} &= \datapointunconstrained{\indexdata} - \bwribr^{-1} \bwribl
	\left( \variableconstrained - \datapointconstrained{\indexdata} \right),
	\forall \indexdata \in \{1,\ldots,\numberofsamples\}.
\end{align}

To generate samples from \cref{eq:gaussian kde} that satisfy \cref{eq:linear constraint}, two random numbers need to be generated. 
First, an integer $\indexsampling\in\{1,\ldots,\numberofsamples\}$ is randomly chosen with the likelihood of the integer $j$ proportional to the weight $\weight{\indexsampling}$ of \cref{eq:weights constrained hard}. 
Next, a random sample is drawn from a Gaussian with covariance $\bwribr^{-1}$ and mean $\datapointunconstrainedtranslated{\indexsampling}$.
Finally, this random sample is mapped according to \cref{eq:rotation variable} to obtain the final random sample.
The procedure for sampling is summarized in \cref{alg:constrained hard}.

\begin{algorithm}[t]
	\SetKwInOut{Input}{Input}
	\SetKwInOut{Output}{Output}
	\Input{$\datapoint{1}, \ldots, \datapoint{\numberofsamples}$, $\constraintmatrix$, $\constraintvector$, $\bandwidthmatrix$}
	\Output{Sample $\variable$ from \cref{eq:gaussian kde} while satisfying $\constraintmatrix \variable = \constraintvector$}
	
	$\svdu$, $\svds$, $\svdva$, $\svdvb$ $\gets$ Perform \iac{svd} of $\constraintmatrix$; see \cref{eq:svd constraint matrix}
	
	$\datapointconstrained{1},\ldots,\datapointconstrained{\numberofsamples}$ $\gets$ Map the data points using $\datapointconstrained{\indexdata} = \svdva\transpose \datapoint{\indexdata}$
		
	$\datapointunconstrained{1},\ldots,\datapointunconstrained{\numberofsamples}$ $\gets$ Map the data points using $\datapointunconstrained{\indexdata} = \svdvb\transpose \datapoint{\indexdata}$
	
	$\variableconstrained$ $\gets$ Compute $\variableconstrained$ using \cref{eq:fixed part}
	
	$\bwriul$, $\bwriur$, $\bwribl$, $\bwribr$, $\bwrischur$ $\gets$ Compute $\svdv\transpose \bandwidthmatrix^{-1} \svdv$ according to \cref{eq:rotated matrix inverse} and $\bwrischur$ according to \cref{eq:rotated schur complement}
	
	$\weight{1},\ldots,\weight{\numberofsamples}$ $\gets$ Compute the weights according to \cref{eq:weights constrained hard}
	
	$\indexsampling$ $\gets$ Generate a random integer $\indexsampling\in\{1,\ldots,\numberofsamples\}$ with the likelihood of $\indexsampling$ proportional to $\weight{\indexsampling}$
	
	$\datapointunconstrainedtranslated{\indexsampling}$ $\gets$ Compute the mean of the Gaussian to generate a sample from according to \cref{eq:offset constrained hard}
	
	$\variableunconstrained$ $\gets$ Generate a random sample from a Gaussian with covariance $\bwribr^{-1}$ and mean $\datapointunconstrainedtranslated{\indexsampling}$
	
	$\variable$ $\gets$ Compute $\variable$ according to \cref{eq:rotation variable}
		
	\caption{Sampling with linear equality constraints and full bandwidth matrix.}
	\label{alg:constrained hard}
\end{algorithm}

\section{Example}
\label{sec:example}

\cstartb To illustrate the proposed method, we have applied it to generate a new set of parameters that describe scenarios.
The \ac{ngsim} data set is used as a data source.
The \ac{ngsim} data set contains vehicles' trajectories obtained from video footage of cameras that were located at several motorways in the US \autocite{kovvali2007video}.
In total, 18182 longitudinal interactions between two vehicles are analyzed. 
In each of these longitudinal interactions, we look at the speed profile of the leading vehicle. 
The speed profile is split into parts of \SI{5}{\second}, resulting in $\numberofsamples=99840$ data samples.

We first apply the method of conditional sampling in a straightforward example to illustrate that \cref{alg:constrained hard} produces correct results.
The second example explains the usefulness of sampling with a linear equality constraint in case a parameter reduction technique is used.

\subsection{Sampling with a linear equality constraint}
\label{sec:straightforward example}

In this first example, 2 parameters describe a scenario:
\begin{enumerate}
	\item The speed of the leading vehicle at a certain time $\time$ and
	\item The speed of the same vehicle at time $\time+\timestep$ with $\timestep=\SI{5}{\second}$.
\end{enumerate}
The bandwidth matrix is estimated using the plug-in selector of \textcite{wand1994multivariate}.
We want to sample the initial speed in case the speed is reduced by \SI{5}{\meter\per\second}. 
To achieve this, we use
\begin{equation*}
	\constraintmatrix = \begin{bmatrix}	1 & -1 \end{bmatrix},
	\constraintvector = \begin{bmatrix} 5 \end{bmatrix}.
\end{equation*} \cendb

In \cref{fig:constrained hard}, the result of \cref{alg:constrained hard} is shown. 
In total, $10^6$ samples are generated and shown by the histogram. 
Because the histogram follows the same pattern as the actual density of the speed difference according to the \ac{kde}, it illustrates that the provided algorithm correctly samples from the \ac{kde}. 

%

\setlength{\figurewidth}{.9\linewidth}
\setlength{\figureheight}{.7\figurewidth}
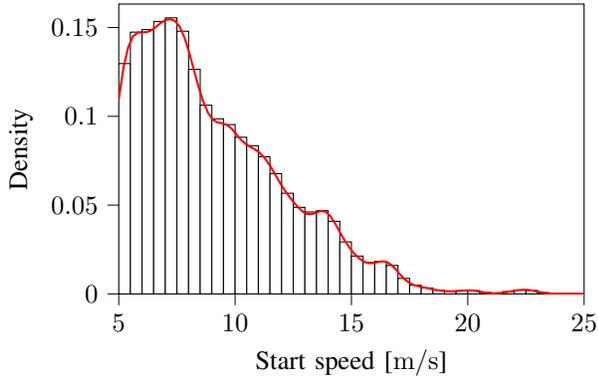
\begin{figure}
	\centering
	\begin{tikzpicture}
	
	\begin{axis}[
	height=\figureheight,
	scaled y ticks=false,
	tick align=outside,
	tick pos=left,
	width=\figurewidth,
	x grid style={white!69.0196078431373!black},
	xlabel={Start speed [\si{\meter\per\second}]},
	xmin=5, xmax=25,
	xtick style={color=black},
	xticklabel style={align=center},
	y grid style={white!69.0196078431373!black},
	ylabel={Density},
	ymin=0, ymax=0.163223142242901,
	ytick style={color=black},
	yticklabel style={/pgf/number format/fixed,/pgf/number format/precision=3}
	]
	\draw[draw=black,fill=white] (axis cs:5,0) rectangle (axis cs:5.5,0.129692616323191);
	\draw[draw=black,fill=white] (axis cs:5.5,0) rectangle (axis cs:6,0.147417753040331);
	\draw[draw=black,fill=white] (axis cs:6,0) rectangle (axis cs:6.5,0.148867895412181);
	\draw[draw=black,fill=white] (axis cs:6.5,0) rectangle (axis cs:7,0.153431827454197);
	\draw[draw=black,fill=white] (axis cs:7,0) rectangle (axis cs:7.5,0.155450611659906);
	\draw[draw=black,fill=white] (axis cs:7.5,0) rectangle (axis cs:8,0.147933547120115);
	\draw[draw=black,fill=white] (axis cs:8,0) rectangle (axis cs:8.5,0.126481586785525);
	\draw[draw=black,fill=white] (axis cs:8.5,0) rectangle (axis cs:9,0.10626837780648);
	\draw[draw=black,fill=white] (axis cs:9,0) rectangle (axis cs:9.5,0.0986308203873952);
	\draw[draw=black,fill=white] (axis cs:9.5,0) rectangle (axis cs:10,0.095350031814348);
	\draw[draw=black,fill=white] (axis cs:10,0) rectangle (axis cs:10.5,0.0882155850140681);
	\draw[draw=black,fill=white] (axis cs:10.5,0) rectangle (axis cs:11,0.0834719706081931);
	\draw[draw=black,fill=white] (axis cs:11,0) rectangle (axis cs:11.5,0.0772570747288382);
	\draw[draw=black,fill=white] (axis cs:11.5,0) rectangle (axis cs:12,0.0677360233544795);
	\draw[draw=black,fill=white] (axis cs:12,0) rectangle (axis cs:12.5,0.0566992983932169);
	\draw[draw=black,fill=white] (axis cs:12.5,0) rectangle (axis cs:13,0.0487573379106535);
	\draw[draw=black,fill=white] (axis cs:13,0) rectangle (axis cs:13.5,0.0461360893084766);
	\draw[draw=black,fill=white] (axis cs:13.5,0) rectangle (axis cs:14,0.0468696161350535);
	\draw[draw=black,fill=white] (axis cs:14,0) rectangle (axis cs:14.5,0.0408703390923293);
	\draw[draw=black,fill=white] (axis cs:14.5,0) rectangle (axis cs:15,0.0292839974886747);
	\draw[draw=black,fill=white] (axis cs:15,0) rectangle (axis cs:15.5,0.0213166700841548);
	\draw[draw=black,fill=white] (axis cs:15.5,0) rectangle (axis cs:16,0.0177948957525203);
	\draw[draw=black,fill=white] (axis cs:16,0) rectangle (axis cs:16.5,0.0183931323286623);
	\draw[draw=black,fill=white] (axis cs:16.5,0) rectangle (axis cs:17,0.0160255529460509);
	\draw[draw=black,fill=white] (axis cs:17,0) rectangle (axis cs:17.5,0.00872622115305344);
	\draw[draw=black,fill=white] (axis cs:17.5,0) rectangle (axis cs:18,0.00485353773435337);
	\draw[draw=black,fill=white] (axis cs:18,0) rectangle (axis cs:18.5,0.00336534497956906);
	\draw[draw=black,fill=white] (axis cs:18.5,0) rectangle (axis cs:19,0.00214561881549158);
	\draw[draw=black,fill=white] (axis cs:19,0) rectangle (axis cs:19.5,0.0015431544190235);
	\draw[draw=black,fill=white] (axis cs:19.5,0) rectangle (axis cs:20,0.00179893754875205);
	\draw[draw=black,fill=white] (axis cs:20,0) rectangle (axis cs:20.5,0.00184967139266515);
	\draw[draw=black,fill=white] (axis cs:20.5,0) rectangle (axis cs:21,0.000796944131468301);
	\draw[draw=black,fill=white] (axis cs:21,0) rectangle (axis cs:21.5,0.000407984661467857);
	\draw[draw=black,fill=white] (axis cs:21.5,0) rectangle (axis cs:22,0.00127257391815363);
	\draw[draw=black,fill=white] (axis cs:22,0) rectangle (axis cs:22.5,0.00214561881549158);
	\draw[draw=black,fill=white] (axis cs:22.5,0) rectangle (axis cs:23,0.00194691126016527);
	\draw[draw=black,fill=white] (axis cs:23,0) rectangle (axis cs:23.5,0.000676451252174685);
	\draw[draw=black,fill=white] (axis cs:23.5,0) rectangle (axis cs:24,9.7239867500111e-05);
	\draw[draw=black,fill=white] (axis cs:24,0) rectangle (axis cs:24.5,6.34173048913767e-06);
	\draw[draw=black,fill=white] (axis cs:24.5,0) rectangle (axis cs:25,1.47973711413212e-05);
	\addplot [thick, red]
	table {%
	5 0.109526920885485
	5.2020202020202 0.127284157932305
	5.4040404040404 0.138871645239083
	5.60606060606061 0.144946087591813
	5.80808080808081 0.147101186471081
	6.01010101010101 0.147304149181364
	6.21212121212121 0.147424686319923
	6.41414141414141 0.14854702691976
	6.61616161616162 0.150584580324225
	6.81818181818182 0.152734025378058
	7.02020202020202 0.154257638714586
	7.22222222222222 0.154713294246049
	7.42424242424242 0.15362160446291
	7.62626262626263 0.1502973712178
	7.82828282828283 0.144256568850154
	8.03030303030303 0.135793304820704
	8.23232323232323 0.126036477802127
	8.43434343434343 0.116413606992252
	8.63636363636364 0.108115443550819
	8.83838383838384 0.101926513869218
	9.04040404040404 0.0981761080755135
	9.24242424242424 0.0965603215192872
	9.44444444444444 0.0960395623806687
	9.64646464646465 0.0952047289240201
	9.84848484848485 0.0931212216421595
	10.0505050505051 0.0899795508164766
	10.2525252525253 0.0867968042609609
	10.4545454545455 0.0843987954095828
	10.6565656565657 0.0827733655976383
	10.8585858585859 0.0813172684509002
	11.0606060606061 0.0793997579884564
	11.2626262626263 0.0766379936130021
	11.4646464646465 0.0729463676722193
	11.6666666666667 0.0685561202720152
	11.8686868686869 0.0639393889282269
	12.0707070707071 0.0595843181991533
	12.2727272727273 0.0557429726286022
	12.4747474747475 0.0523489586429014
	12.6767676767677 0.0492577604353637
	12.8787878787879 0.0466471013245834
	13.0808080808081 0.0450642613446621
	13.2828282828283 0.0449030190341437
	13.4848484848485 0.0458135007410698
	13.6868686868687 0.046683252922713
	13.8888888888889 0.0462481111178903
	14.0909090909091 0.0438535941445535
	14.2929292929293 0.039798990580635
	14.4949494949495 0.0350153513883531
	14.6969696969697 0.0303914603136225
	14.8989898989899 0.026361068751994
	15.1010101010101 0.0229959178514887
	15.3030303030303 0.0203025575903587
	15.5050505050505 0.0183793129488047
	15.7070707070707 0.017373294361514
	15.9090909090909 0.0173068930486991
	16.1111111111111 0.0178696745381315
	16.3131313131313 0.0183679216929493
	16.5151515151515 0.0180009111212062
	16.7171717171717 0.0163382442442824
	16.9191919191919 0.0136043789664418
	17.1212121212121 0.0105147973517004
	17.3232323232323 0.00782156093628824
	17.5252525252525 0.00593142775080837
	17.7272727272727 0.00480908558488148
	17.9292929292929 0.00414080094666354
	18.1313131313131 0.00360086946699105
	18.3333333333333 0.00303603840697512
	18.5353535353535 0.00247047025397191
	18.7373737373737 0.00199225241265001
	18.9393939393939 0.00165828755991812
	19.1414141414141 0.00148170922009913
	19.3434343434343 0.00146046208195002
	19.5454545454545 0.00158331690380024
	19.7474747474747 0.00179645862882738
	19.9494949494949 0.0019733345691334
	20.1515151515151 0.00195583531870928
	20.3535353535354 0.00166838664408713
	20.5555555555556 0.00119452230870342
	20.7575757575758 0.000720785422078628
	20.959595959596 0.000408905185680007
	21.1616161616162 0.000328728613992972
	21.3636363636364 0.000477891217058044
	21.5656565656566 0.000813835936459417
	21.7676767676768 0.00126034819589294
	21.969696969697 0.00171844577654117
	22.1717171717172 0.00209065399716433
	22.3737373737374 0.00228707206715084
	22.5757575757576 0.00222661475630252
	22.7777777777778 0.00188485099388466
	22.979797979798 0.00135320531443892
	23.1818181818182 0.000808069053632275
	23.3838383838384 0.00039623746038018
	23.5858585858586 0.000158323682609431
	23.7878787878788 5.13536712896669e-05
	23.989898989899 1.36553957184634e-05
	24.1919191919192 3.65745663859596e-06
	24.3939393939394 3.14468320656596e-06
	24.5959595959596 7.12256406819267e-06
	24.7979797979798 1.50840076320442e-05
	25 2.59239470606882e-05
	};
	\end{axis}
	
	\end{tikzpicture}
	
	\caption{The histogram shows the result of the conditional sampling according to \cref{alg:constrained hard}. The red line represents the true \ac{pdf}.}
	\label{fig:constrained hard}
\end{figure}

\cstartb
\subsection{Applying conditional sampling with parameter reduction}
\label{sec:example parameter reduction}

Typically, more than 2 parameters are needed to describe a scenario. 
If the number of parameters is too high, however, the \ac{pdf} estimation using the \ac{kde} suffers from the curse of dimensionality \autocite{scott1992multivariate}.
To avoid the curse of dimensionality, a reduction of parameters can be obtained.
In this example, we use \iac{svd} to reduce the number of parameters.

Instead of using the speed at only two time instances, we consider the following scenario parameters:
\begin{equation}
	\label{eq:parameters large}
	\datapoint{\indexdata} = \begin{bmatrix}
		\speed{\timedatapoint{\indexdata}} \\
		\speed{\timedatapoint{\indexdata} + \timestep} \\
		\vdots \\
		\speed{\timedatapoint{\indexdata} + \timestepnumberof\timestep}
	\end{bmatrix} \in \realnumbers^{\timestepnumberof+1},
\end{equation}
where $\speed{\cdot}$ denotes the speed of the leading vehicle, $\timedatapoint{\indexdata}$ denotes the time of the $\indexdata$-th data point, $\timestep$ denotes the time step, and $\timestepnumberof$ denote the number of time steps.
We used $\timestep=\SI{0.1}{\second}$ and $\timestepnumberof=50$, so this results in 51 parameters.
To reduce these 51 parameters to $\dimension$ parameters, consider the following matrix:
\begin{equation*}
	\datapointmatrix = \begin{bmatrix}
		\datapoint{1}-\datapointmean & \cdots & \datapoint{\numberofsamples}-\datapointmean
	\end{bmatrix} \in \realnumbers^{(\timestepnumberof+1) \times \numberofsamples},
\end{equation*}
with $\datapointmean=\frac{1}{\numberofsamples}\sum_{\indexdata=1}^{\numberofsamples} \datapoint{\indexdata}$ and the following \ac{svd} of this matrix:
\begin{equation*}
	\datapointmatrix =
	\begin{bmatrix} \svduBa & \svduBb \end{bmatrix}
	\begin{bmatrix} \svdsBa & 0 \\ 0 & \svdsBb \end{bmatrix}
	\begin{bmatrix} \svdvBa\transpose \\ \svdvBb\transpose \end{bmatrix} \approx
	\svduBa \svdsBa \svdvBa\transpose,
\end{equation*}
with $\svduBa\in\realnumbers^{(\timestepnumberof+1)\times\dimension}$, $\svduBb\in\realnumbers^{(\timestepnumberof+1)\times(\timestepnumberof+1-\dimension)}$, $\svdsBa\in\realnumbers^{\dimension\times\dimension}$, $\svdsBb\in\realnumbers^{(\timestepnumberof+1-\dimension)\times(\numberofsamples-\dimension)}$, $\svdvBa\in\realnumbers^{\numberofsamples\times\dimension}$, and $\svdvBb\in\realnumbers^{\numberofsamples\times(\numberofsamples-\dimension)}$. 
Using this approximation, it follows that
\begin{equation}
	\label{eq:parameter reduction}
	\datapoint{\indexdata} \approx \svduBa \svdsBa \svdvBvec{\indexdata} + \datapointmean, 
	\forall \indexdata \in \{1, \ldots, \numberofsamples\},
\end{equation}
where $\svdvBvec{\indexdata}\in\realnumbers^{\dimension}$ is the $\indexdata$-th column of $\svdvBa\transpose$.

Instead of using the original data points $\datapoint{\indexdata}$, the vectors $\svdvBvec{\indexdata}$ are used for the estimation of the \ac{kde}.
For each sample of this \ac{kde}, the mapping of \cref{eq:parameter reduction} is then applied to obtain the scenario parameters according to \cref{eq:parameters large}.
In this example, $\dimension=4$ is used.
As with the first example, the bandwidth matrix is estimated using the plug-in selector of \textcite{wand1994multivariate}.

\begin{figure}
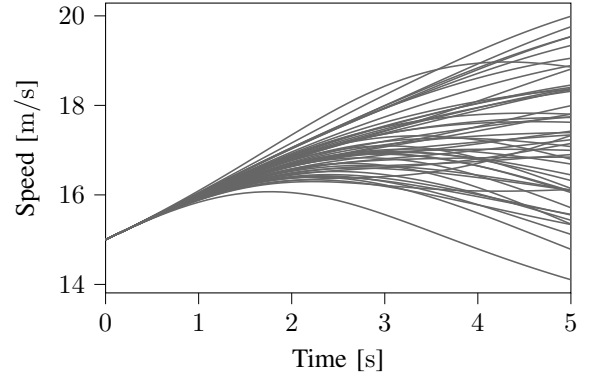
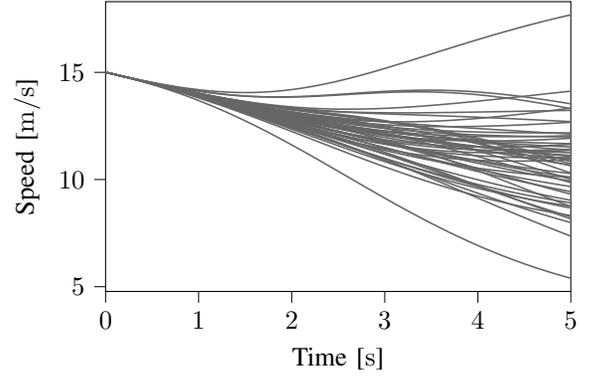

	\centering
	\begin{subfigure}{\linewidth}
		\centering

		
		\caption{$\speedinit=\SI{15}{\meter\per\second}$ and $\accelerationinit=\SI{1}{\meter\per\second\squared}$.}
		\label{fig:profiles init acceleration}
		\vspace{1em}
	\end{subfigure}
	\begin{subfigure}{\linewidth}
		\centering
		%
		
		\caption{$\speedinit=\SI{15}{\meter\per\second}$ and $\accelerationinit=\SI{-1}{\meter\per\second\squared}$.}
		\label{fig:profiles init deceleration}
	\end{subfigure}
	\caption{50 scenarios sampled from the \ac{kde} with a constraint on the initial speed and the initial acceleration.}
	\label{fig:profiles init speed acceleration}
\end{figure}

In \cref{fig:profiles init speed acceleration}, 50 scenarios are sampled from the \ac{kde} in which the initial speed $\speedinit$ and initial acceleration $\accelerationinit$ are fixed by a linear equality constraint.
To achieve this, the following constraint matrix and constraint vector are used:
\begin{equation*}
	\constraintmatrix=\begin{bmatrix}\svduBvec{1} \\ \svduBvec{2} \end{bmatrix} \svdsBa,
	\constraintvector=\begin{bmatrix}\speedinit-\datapointmeanentry{1} \\ \speedinit+\timestep\cdot\accelerationinit-\datapointmeanentry{2} \end{bmatrix},
\end{equation*}
where $\svduBvec{1}$ and $\svduBvec{2}$ denote the first and second row of $\svduBa$, respectively, and $\datapointmeanentry{1}$ and $\datapointmeanentry{2}$ denote the first and second entry of $\datapointmean$, respectively.
\Cref{fig:profiles init acceleration} shows the result with $\speedinit=\SI{15}{\meter\per\second}$ and $\accelerationinit=\SI{1}{\meter\per\second\squared}$.
In \cref{fig:profiles init deceleration}, $\accelerationinit=\SI{-1}{\meter\per\second\squared}$ is used instead.

\setlength{\figurewidth}{.9\linewidth}
\setlength{\figureheight}{.7\figurewidth}

\begin{figure}
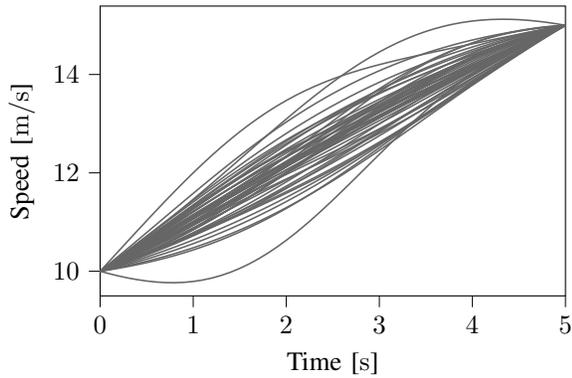
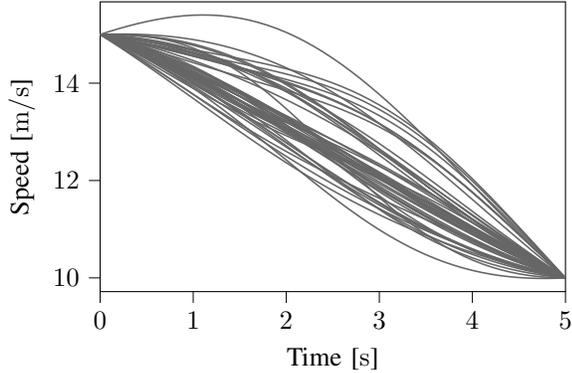

	\centering
	\begin{subfigure}{\linewidth}
		\centering

		
		\caption{$\speedinit=\SI{10}{\meter\per\second}$ and $\speedend=\SI{15}{\meter\per\second}$.}
		\label{fig:profiles accelerating}
		\vspace{1em}
	\end{subfigure}
	\begin{subfigure}{\linewidth}
		\centering
		%
		
		\caption{$\speedinit=\SI{15}{\meter\per\second}$ and $\speedend=\SI{10}{\meter\per\second}$.}
		\label{fig:profiles decelerating}
	\end{subfigure}
	\caption{50 scenarios sampled from the \ac{kde} with a constraint on the initial speed and the end speed.}
	\label{fig:profiles accelerating decelerating}
\end{figure}

In \cref{fig:profiles accelerating decelerating}, 50 scenarios are sampled from the \ac{kde} in which the initial speed $\speedinit$ and end speed $\speedend$ are fixed by a linear equality constraint.
To achieve this, the following constraint matrix and constraint vector are used:
\begin{equation*}
	\constraintmatrix=\begin{bmatrix}\svduBvec{1} \\ \svduBvec{\timestepnumberof+1} \end{bmatrix} \svdsBa,
	\constraintvector=\begin{bmatrix}\speedinit-\datapointmeanentry{1} \\ \speedend-\datapointmeanentry{\timestepnumberof+1} \end{bmatrix}.
\end{equation*}
This illustrates that the conditional sampling can be used to generated scenarios in which a leading vehicle is accelerating (\cref{fig:profiles accelerating}) or decelerating (\cref{fig:profiles decelerating}).
\Cref{fig:profiles accelerating} shows the result with $\speedinit=\SI{10}{\meter\per\second}$ and $\speedend=\SI{5}{\meter\per\second}$.
In \cref{fig:profiles decelerating}, the start and end speed are opposite: $\speedinit=\SI{15}{\meter\per\second}$ and $\speedend=\SI{10}{\meter\per\second}$.

Note that all speed profiles in \cref{fig:profiles init speed acceleration,fig:profiles accelerating decelerating} are all drawn from the same \ac{kde}.
The only difference between these scenarios is that different conditions are used.
\cendb

\section{Discussion}
\label{sec:discussion}

\cstartb This paper has provided a method for sampling from \iac{kde} such that the samples satisfy a linear equality constraint.
The provided algorithm calculates weights that are used to weigh the contribution of each input data point to the overall \ac{pdf}.
Samples are drawn by randomly picking a random data point with the likelihood proportional to the calculated weights and adding an offset to this data point using a zero-mean multivariate Gaussian with a covariance matrix that equals the original bandwidth matrix (pre)multiplied with a rotation matrix. 

The provided algorithm can be regarded as a generalization of the conditional sampling in \autocite{holmes2007fast}.
With conditional sampling, few parameters are fixed. 
This is the same as considering the linear equality constraint of \cref{eq:linear constraint} with 
\begin{equation*}
	\constraintmatrix = \begin{bmatrix} \identitymatrix{\numberofconstraints} & 0 \end{bmatrix}.
\end{equation*}
In this particular case, the rotation of the data is not needed, so steps 2, 3, 4, and 10 of \cref{alg:constrained hard} can be skipped.
Note that this way of sampling becomes impractical if a parameter reduction is used as shown in \cref{sec:example parameter reduction}, because the set of reduced parameters have no physical meaning anymore. \cendb

The computational cost of the provided algorithm scales quadratically with respect to number of parameters that are used to describe a single scenario ($\dimension$) and linearly with the number of data points ($\numberofsamples$).
Because the number of data points is generally much larger than the dimension of the data points, let us assume that $\numberofsamples \gg \dimension$.
Looking at \cref{alg:constrained hard} and considering $\numberofsamples \gg \dimension$, steps 2, 3, and 6 are the most time consuming, because these steps contain a loop over the data points. 
It is easy to see that the number of computations of these steps scales linearly with $\numberofsamples$. 
Since these computations contain a multiplication of a $\dimension$-by-$\dimension$ matrix and a vector with $\dimension$ rows, the computational cost scales quadratically with $\dimension$. 

If we want to sample multiple times using the same linear constraint, it suffices to perform steps 1 till 6 of \cref{alg:constrained hard} only once.
Step 7 of \cref{alg:constrained hard} does not depend on $\dimension$ and scales linearly with $\numberofsamples$ \autocite{vose1991linear}.
Steps 8 till 10 of \cref{alg:constrained hard} do not depend on $\numberofsamples$ and scale quadratically with $\dimension$.
Because these steps do not depend on $\numberofsamples$ and $\numberofsamples \gg \dimension$, the computational cost of these steps is minor compared to step 7 of \cref{alg:constrained hard}. 
\cstarta Therefore, if we want to draw many samples, i.e., more than $\numberofsamples$, the computational cost is dominated by the computational cost of step 7, which means that, in that case, the computational cost scales linearly with $\numberofsamples$. \cenda

\cstartb Another application of the conditional sampling is to predict how the future will develop based on some initial conditions. 
For example, \cref{fig:profiles init acceleration,fig:profiles init deceleration} each show 50 possibilities for the speed of the leading vehicle in the next \SI{5}{\second} given an initial speed and an initial acceleration. 
This could be used, for instance, to determine real-time a worst-case scenario, such that \iac{av} could proactively respond to such a scenario. 
\cstartc Similarly, when using Bayesian networks for predicting continuous variables \autocite{bishop2006pattern}, our algorithm provides a way to sample from the conditional densities. \cendc

Note that for an efficient scenario-based assessment of \iac{av}, scenarios that might lead to critical behavior need to be emphasized. 
\cstartc Several techniques are proposed in literature to emphasize these scenarios, such as reachability analysis \autocite{althoff2018automatic}, boundary searching \autocite{zhou2018safety}, and importance sampling \autocite{zhao2018evaluation, deGelder2017assessment, xu2018accelerated, jesenski2020scalable}. \cendc
With importance sampling, a different \ac{pdf} is used to sample scenario parameters, such that more emphasis is put on scenarios that might lead to critical behavior.
Our proposed method for conditional sampling can be combined with the importance sampling techniques explained in \autocite{deGelder2017assessment, xu2018accelerated, jesenski2020scalable}. \cendb


\cstartb This work comes with limitations. \cendb
\cstarta It should be emphasized that the method only works when using a Gaussian kernel for the \ac{kde}. 
In practice, this is usually not a problem, because the choice of the kernel is not crucial \autocite{duong2007ks}. \cenda
\cstartb Another limitation is that our method cannot be extended to deal with (linear) inequality constraints. 
If these inequality constraints are not too severe, however, straightforward rejection sampling could be used in that case, i.e., sample data points until a data point satisfies the linear inequality constraint. 
It should also be noted that in practice, more parameters for describing a scenario might need to be considered. 
For example, the state of neighboring vehicles (instead of only the leading vehicle), lane curvature, etc.
Although these parameters are not considered in the example of this paper, a parameter reduction technique as explained in \cref{sec:example parameter reduction} can still be useful in case these parameters are considered. \cendb

\acresetall
\section{Conclusions}
\label{sec:conclusions}

It is expected that scenario-based test descriptions become more and more important for the assessment of \acp{av}. 
One way to generate these scenario-based test descriptions is to sample the scenario parameters from \iac{pdf}.
To deal with the unknown shape of the \ac{pdf}, it is proposed to estimate the \ac{pdf} using \ac{kde}. 
In this paper, we have shown how these parameters can be drawn from the estimated \ac{pdf} while these parameters are subject to a linear equality constraint. 
Through an example, we have illustrated the effectiveness of our method by generating different scenarios of a longitudinal interaction with a leading vehicle. 

Future work involves applying this method using more complex scenarios, e.g., scenarios that contain several different actors, to generate scenario-based test cases for the safety assessment of \acp{av}.


\bibliographystyle{ieeetran}
\bibliography{references}

\begin{thebibliography}{10}
\providecommand{\url}[1]{#1}
\csname url@samestyle\endcsname
\providecommand{\newblock}{\relax}
\providecommand{\bibinfo}[2]{#2}
\providecommand{\BIBentrySTDinterwordspacing}{\spaceskip=0pt\relax}
\providecommand{\BIBentryALTinterwordstretchfactor}{4}
\providecommand{\BIBentryALTinterwordspacing}{\spaceskip=\fontdimen2\font plus
\BIBentryALTinterwordstretchfactor\fontdimen3\font minus
  \fontdimen4\font\relax}
\providecommand{\BIBforeignlanguage}[2]{{%
\expandafter\ifx\csname l@#1\endcsname\relax
\typeout{** WARNING: IEEEtran.bst: No hyphenation pattern has been}%
\typeout{** loaded for the language `#1'. Using the pattern for}%
\typeout{** the default language instead.}%
\else
\language=\csname l@#1\endcsname
\fi
#2}}
\providecommand{\BIBdecl}{\relax}
\BIBdecl

\bibitem{bengler2014threedecades}
K.~Bengler, K.~Dietmayer, B.~F{\"a}rber, M.~Maurer, C.~Stiller, and H.~Winner,
  ``Three decades of driver assistance systems: Review and future
  perspectives,'' \emph{IEEE Intelligent Transportation Systems Magazine},
  vol.~6, no.~4, pp. 6--22, 2014.

\bibitem{stellet2015taxonomy}
J.~E. Stellet, M.~R. Zofka, J.~Schumacher, T.~Schamm, F.~Niewels, and J.~M.
  Z\"{o}llner, ``Testing of advanced driver assistance towards automated
  driving: A survey and taxonomy on existing approaches and open questions,''
  in \emph{IEEE 18th International Conference on Intelligent Transportation
  Systems}, 9 2015, pp. 1455--1462.

\bibitem{koopman2016challenges}
P.~Koopman and M.~Wagner, ``Challenges in autonomous vehicle testing and
  validation,'' \emph{SAE International Journal of Transportation Safety},
  vol.~4, pp. 15--24, 2016.

\bibitem{kalra2016driving}
N.~Kalra and S.~M. Paddock, ``Driving to safety: How many miles of driving
  would it take to demonstrate autonomous vehicle reliability?''
  \emph{Transportation Research Part A: Policy and Practice}, vol.~94, pp.
  182--193, 2016.

\bibitem{zhao2018evaluation}
D.~Zhao, X.~Huang, H.~Peng, H.~Lam, and D.~J. LeBlanc, ``Accelerated evaluation
  of automated vehicles in car-following maneuvers,'' \emph{IEEE Transactions
  on Intelligent Transportation Systems}, vol.~19, no.~3, pp. 733--744, 3 2018.

\bibitem{riedmaier2020survey}
S.~Riedmaier, T.~Ponn, D.~Ludwig, B.~Schick, and F.~Diermeyer, ``Survey on
  scenario-based safety assessment of automated vehicles,'' \emph{IEEE Access},
  vol.~8, pp. 87\,456--87\,477, 2020.

\bibitem{elrofai2018scenario}
\BIBentryALTinterwordspacing
H.~Elrofai, J.-P. Paardekooper, E.~de~Gelder, S.~Kalisvaart, and O.~Op~den
  Camp, ``Scenario-based safety validation of connected and automated
  driving,'' Netherlands Organization for Applied Scientific Research, TNO,
  Tech. Rep., 2018. [Online]. Available:
  \url{http://publications.tno.nl/publication/34626550/AyT8Zc/TNO-2018-streetwise.pdf}
\BIBentrySTDinterwordspacing

\bibitem{putz2017pegasus}
A.~P\"{u}tz, A.~Zlocki, J.~Bock, and L.~Eckstein, ``System validation of highly
  automated vehicles with a database of relevant traffic scenarios,'' in
  \emph{12th ITS European Congress}, 2017, pp. 1--8.

\bibitem{deGelder2017assessment}
E.~de~Gelder and J.-P. Paardekooper, ``Assessment of automated driving systems
  using real-life scenarios,'' in \emph{IEEE Intelligent Vehicles Symposium
  (IV)}, 2017, pp. 589--594.

\bibitem{jacobo2019development}
J.~Antona-Makoshi, N.~Uchida, K.~Yamazaki, K.~Ozawa, E.~Kitahara, and
  S.~Taniguchi, ``Development of a safety assurance process for autonomous
  vehicles in {J}apan,'' in \emph{26th International Technical Conference on
  the Enhanced Safety of Vehicles (ESV)}, 2019, pp. 1--18.

\bibitem{krajewski2018highD}
R.~Krajewski, J.~Bock, L.~Kloeker, and L.~Eckstein, ``The high{D} dataset: A
  drone dataset of naturalistic vehicle trajectories on {G}erman highways for
  validation of highly automated driving systems,'' in \emph{IEEE 21st
  International Conference on Intelligent Transportations Systems (ITSC)},
  2018, pp. 2118--2125.

\bibitem{xu2018accelerated}
Y.~Xu, Y.~Zou, and J.~Sun, ``Accelerated testing for automated vehicles safety
  evaluation in cut-in scenarios based on importance sampling, genetic
  algorithm and simulation applications,'' \emph{Journal of Intelligent and
  Connected Vehicles}, vol.~1, pp. 28--38, 2018.

\bibitem{jesenski2020scalable}
S.~Jesenski, N.~Tiemann, J.~E. Stellet, and J.~M. Z{\"o}llner, ``Scalable
  generation of statistical evidence for the safety of automated vehicles by
  the use of importance sampling,'' in \emph{IEEE 23rd International Conference
  on Intelligent Transportation Systems (ITSC)}.\hskip 1em plus 0.5em minus
  0.4em\relax IEEE, 2020, pp. 1--8.

\bibitem{degelder2019risk}
E.~de~Gelder, A.~Khabbaz~Saberi, and H.~Elrofai, ``A method for scenario risk
  quantification for automated driving systems,'' in \emph{26th International
  Technical Conference on the Enhanced Safety of Vehicles (ESV)}, 2019.

\bibitem{parzen1962estimation}
E.~Parzen, ``On estimation of a probability density function and mode,''
  \emph{The Annals of Mathematical Statistics}, vol.~33, no.~3, pp. 1065--1076,
  1962.

\bibitem{rosenblatt1956remarks}
M.~Rosenblatt, ``Remarks on some nonparametric estimates of a density
  function,'' \emph{The Annals of Mathematical Statistics}, vol.~27, no.~3, pp.
  832--837, 1956.

\bibitem{hall1999density}
P.~Hall and B.~Presnell, ``Density estimation under constraints,''
  \emph{Journal of Computational and Graphical Statistics}, vol.~8, no.~2, pp.
  259--277, 1999.

\bibitem{wolters2018practical}
M.~A. Wolters and W.~J. Braun, ``A practical implementation of weighted kernel
  density estimation for handling shape constraints,'' \emph{Stat}, vol.~7,
  no.~1, p. e202, 2018.

\bibitem{holmes2007fast}
M.~P. Holmes, A.~G. Gray, and C.~L. Isbell, ``Fast nonparametric conditional
  density estimation,'' in \emph{23rd Conference on Uncertainty in Artificial
  Intelligence}, 2007, pp. 175--182.

\bibitem{turlach1993bandwidthselection}
B.~A. Turlach, ``Bandwidth selection in kernel density estimation: A review,''
  Institut f{\"u}r Statistik und {\"O}konometrie, Humboldt-Universit{\"a}t zu
  Berlin, Tech. Rep., 1993.

\bibitem{duong2007ks}
T.~Duong, ``{ks}: Kernel density estimation and kernel discriminant analysis
  for multivariate data in {R},'' \emph{Journal of Statistical Software},
  vol.~21, no.~7, pp. 1--16, 2007.

\bibitem{silverman1986density}
B.~W. Silverman, \emph{Density Estimation for Statistics and Data
  Analysis}.\hskip 1em plus 0.5em minus 0.4em\relax CRC press, 1986.

\bibitem{jones1996brief}
M.~C. Jones, J.~S. Marron, and S.~J. Sheather, ``A brief survey of bandwidth
  selection for density estimation,'' \emph{Journal of the American Statistical
  Association}, vol.~91, no. 433, pp. 401--407, 1996.

\bibitem{gramacki2017fft}
A.~Gramacki and J.~Gramacki, ``{FFT}-based fast bandwidth selector for
  multivariate kernel density estimation,'' \emph{Computational Statistics \&
  Data Analysis}, vol. 106, pp. 27--45, 2017.

\bibitem{golub2013matrix}
G.~H. Golub and C.~F. Van~Loan, \emph{Matrix Computations}.\hskip 1em plus
  0.5em minus 0.4em\relax John Hopkins University Press, 2013, vol.~3.

\bibitem{zhang2006schur}
F.~Zhang, \emph{The {S}chur Complement and Its Applications}.\hskip 1em plus
  0.5em minus 0.4em\relax Springer Science \& Business Media, 2006, vol.~4.

\bibitem{kovvali2007video}
V.~G. Kovvali, V.~Alexiadis, and L.~Zhang, ``Video-based vehicle trajectory
  data collection,'' in \emph{Transportation Research Board 86th Annual
  Meeting}, 2007.

\bibitem{wand1994multivariate}
M.~P. Wand and M.~C. Jones, ``Multivariate plug-in bandwidth selection,''
  \emph{Computational Statistics}, vol.~9, no.~2, pp. 97--116, 1994.

\bibitem{scott1992multivariate}
D.~W. Scott, \emph{Multivariate Density Estimation: Theory, Practice, and
  Visualization}.\hskip 1em plus 0.5em minus 0.4em\relax John Wiley \& Sons,
  1992.

\bibitem{vose1991linear}
M.~D. Vose, ``A linear algorithm for generating random numbers with a given
  distribution,'' \emph{IEEE Transactions on Software Engineering}, vol.~17,
  no.~9, pp. 972--975, 1991.

\bibitem{bishop2006pattern}
C.~M. Bishop, \emph{Pattern Recognition and Machine Learning}.\hskip 1em plus
  0.5em minus 0.4em\relax Springer, 2006.

\bibitem{althoff2018automatic}
M.~Althoff and S.~Lutz, ``Automatic generation of safety-critical test
  scenarios for collision avoidance of road vehicles,'' in \emph{IEEE
  Intelligent Vehicles Symposium (IV)}.\hskip 1em plus 0.5em minus 0.4em\relax
  IEEE, 2018, pp. 1326--1333.

\bibitem{zhou2018safety}
J.~Zhou and L.~del Re, ``Safety verification of {ADAS} by collision-free
  boundary searching of a parameterized catalog,'' in \emph{Annual American
  Control Conference (ACC)}.\hskip 1em plus 0.5em minus 0.4em\relax IEEE, 2018,
  pp. 4790--4795.

\end{thebibliography}

\end{document}